\documentclass[letterpaper]{article} % DO NOT CHANGE THIS
\usepackage{aaai2026}  % DO NOT CHANGE THIS
\usepackage{times}  % DO NOT CHANGE THIS
\usepackage{helvet}  % DO NOT CHANGE THIS
\usepackage{courier}  % DO NOT CHANGE THIS
\usepackage[hyphens]{url}  % DO NOT CHANGE THIS
\usepackage{graphicx} % DO NOT CHANGE THIS
\usepackage{natbib}  % DO NOT CHANGE THIS AND DO NOT ADD ANY OPTIONS TO IT
\usepackage{caption} % DO NOT CHANGE THIS AND DO NOT ADD ANY OPTIONS TO IT
\usepackage{booktabs}
\usepackage{multirow}
\usepackage{array}
\usepackage{makecell}
\usepackage{amsmath} 
\usepackage{enumitem}
\usepackage{siunitx}
\usepackage{array} 
\usepackage{algorithm}
\usepackage{algorithmic}

\usepackage{newfloat}
\usepackage{listings}
\DeclareCaptionStyle{ruled}{labelfont=normalfont,labelsep=colon,strut=off} % DO NOT CHANGE THIS
\floatstyle{ruled}
\newfloat{listing}{tb}{lst}{}
\floatname{listing}{Listing}
\title{ChipMind: Retrieval-Augmented Reasoning for Long-Context \\Circuit Design Specifications}
\author{
    Changwen Xing\textsuperscript{\rm 1, 2},
    SamZaak Wong\textsuperscript{\rm 1, 2},
    Xinlai Wan\textsuperscript{\rm 2},
    Yanfeng Lu\textsuperscript{\rm 2}, 
    Mengli Zhang\textsuperscript{\rm 2}, 
    Zebin Ma\textsuperscript{\rm 2}, \\
    Lei Qi\textsuperscript{\rm 2, 3},
    Zhengxiong Li\textsuperscript{\rm 4},
    Nan Guan\textsuperscript{\rm 5}, 
    Zhe Jiang\textsuperscript{\rm 1, 2}, 
    Xi Wang\thanks{Corresponding author: xi.wang@seu.edu.cn}\textsuperscript{\rm 1, 2}, 
    Jun Yang\textsuperscript{\rm 1, 2}
    }

\affiliations{
    \textsuperscript{\rm 1}School of Integrated Circuits, Southeast University, Nanjing, China\\
    \textsuperscript{\rm 2}National Center of Technology Innovation for EDA, Nanjing, China\\
    \textsuperscript{\rm 3}School of Computer Science and Engineering, Southeast University, Nanjing, China\\
    \textsuperscript{\rm 4}Department of Computer Science and Engineering, University of Colorado Denver, USA\\
    \textsuperscript{\rm 5}Department of Computer Science, City University of Hong Kong, Hong Kong SAR, China
}

\usepackage{bibentry}
\begin{document}

\maketitle

\begin{abstract}
While Large Language Models (LLMs) demonstrate immense potential for automating integrated circuit (IC) development, their practical deployment is fundamentally limited by restricted context windows. Existing context-extension methods struggle to achieve effective semantic modeling and thorough multi-hop reasoning over extensive, intricate circuit specifications. To address this, we introduce \textbf{ChipMind}, a novel knowledge graph-augmented reasoning framework specifically designed for lengthy IC specifications. ChipMind first transforms circuit specifications into a domain-specific knowledge graph (\textit{ChipKG}) through the \textit{Circuit Semantic-Aware Knowledge Graph Construction} methodology. It then leverages the \textit{ChipKG-Augmented Reasoning} mechanism, combining information-theoretic adaptive retrieval to dynamically trace logical dependencies with intent-aware semantic filtering to prune irrelevant noise, effectively balancing retrieval completeness and precision. Evaluated on an industrial-scale specification reasoning benchmark, ChipMind significantly outperforms state-of-the-art baselines, achieving an average improvement of \textbf{34.59\%} (up to \textbf{72.73\%}). Our framework bridges a critical gap between academic research and practical industrial deployment of LLM-aided Hardware Design (LAD). 
\end{abstract}

% Uncomment the following to link to your code, datasets, an extended version or similar.
% You must keep this block between (not within) the abstract and the main body of the paper.
% \begin{links}
%     \link{Code}{https://aaai.org/example/code}
%     \link{Datasets}{https://aaai.org/example/datasets}
%     \link{Extended version}{https://aaai.org/example/extended-version}
% \end{links}

\section{Introduction}

As modern integrated circuits (ICs) grow exponentially in scale and complexity, traditional human-driven development has become a critical bottleneck in semiconductor design~\cite{itrs2001design,foster2003abv,9218553}. Large Language Model-Aided Hardware Design (LAD) has emerged as a transformative paradigm~\cite{vaswani2017attention,koroteev2021bert}, offering new pathways for advancing Electronic Design Automation (EDA) by generating code and testbenches~\cite{bhandari2024llm} directly from natural language specifications.

\begin{figure}[t]
\centering
\includegraphics[width=\columnwidth]{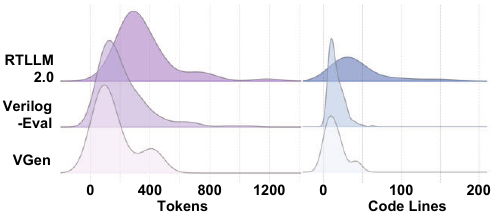} 
\caption{Probability density of specification tokens and design code lines in public LAD benchmarks}
\label{fig:Benchmark_eval}
\end{figure}

However, a fundamental obstacle hinders the adoption of LAD technology in real-world industrial applications: the limited context window of LLMs~\cite{vaswani2017attention}. Our analysis reveals that prominent LAD benchmarks (e.g., VerilogEval~\cite{liu2023verilogeval}, RTLLM2.0~\cite{liu2024openllm}, VGen~\cite{thakur2023benchmarking}) operate on inputs of under 1,000 tokens (Figure \ref{fig:Benchmark_eval}), significantly smaller than actual industrial specifications---ARM's AMBA APB Protocol~\cite{ARM_APB_Protocol_SPEC_2023} contains 7.2k tokens, NXP's I2C-bus specification~\cite{NXP_UM10204_Rev7_2021} spans 49k tokens, and a complex CPU core like the Xuantie C910~\cite{XuanTie_OpenC910_UserManual_2021} demands a staggering 195k tokens. This orders-of-magnitude gap severely restricts current research scalability to realistic, complex industrial scenarios.

To address this challenge, two dominant paradigms exist for extending the context window of LLMs: monolithic long-context models, which leverage techniques like RoPE~\cite{ding2024longrope} and fine-tuning~\cite{chen2023longlora}; and Retrieval-Augmented Generation (RAG)~\cite{wang2024chatcpu} methods. However, both approaches commonly suffer from the ``lost-in-the-middle'' phenomenon~\cite{an2024make}: LLMs tend to over-rely on local context while neglecting the overall document structure and cross-module logical connections. This issue is particularly acute in chip design, since the very nature of a chip specification is a tightly-coupled logical chain that spans the entire document. Consequently, any fragmented understanding leads to significant inaccuracies or failures in downstream tasks.

Recently, Knowledge Graph (KG)-based RAG~\cite{pan2024unifying} has been explored to enhance global context perception by LLMs. However, we identify two fundamental limitations when applying these approaches to IC technical documentation:
\textbf{1) Insufficient Semantic Modeling Capability for Complex IC Documentation:} 
General-purpose KG construction techniques lack the domain-specific semantic precision required to accurately and comprehensively capture intricate interrelations of entities within IC documents. An incomplete and imprecise KG leads to reduced retrieval accuracy, thereby impairing subsequent reasoning performance.
\textbf{2) Retrieval Completeness Bottleneck in Multi-Hop Reasoning:} Tracing chip logic often requires following signal chains across modules, demanding complete contextual coverage. However, the commonly used fixed Top-\texttt{K} retrieval mechanism lacks the adaptability to support intermediate reasoning steps. As a result, critical information is often missed, a failure mode that becomes particularly pronounced in complex multi-hop reasoning tasks.

Ultimately, the core bottleneck in LAD has shifted from how to generate code to how to enable LLMs to perform deep comprehension and reasoning over vast specifications.

To overcome this core bottleneck, we introduce \textbf{ChipMind}: a knowledge graph-augmented reasoning framework tailored for long IC specifications. ChipMind explicitly parses intricate entity-semantic relationships within chip specifications, reconstructing them into a domain-specific knowledge graph (ChipKG), and employs a \textit{Graph-Augmented Reasoning} mechanism with adaptive retrieval. This enables LLMs to iteratively query ChipKG, emulating human experts to accurately explore and verify deep dependency paths. Comprehensive experiments on a newly introduced benchmark for industrial-scale specification reasoning demonstrate significant and consistent advantages of ChipMind, achieving an average improvement of 34.59\% over state-of-the-art baselines, with a maximum performance gain of 72.73\% compared to GraphRAG~\cite{edge2024local}. Our primary contributions are as follows:

\begin{enumerate}
    \item We are the first to systematically identify and address the core reasoning bottleneck for LLMs in industrial chip design. To this end, we propose \textbf{ChipMind}, a novel framework that integrates a domain-specific knowledge graph with multi-hop reasoning. 
    % For broader adoption and reproducibility, ChipMind is publicly available via an interactive web interface (see Appendix A for details).
    
    \item We design a novel \textbf{Circuit Semantic-Aware Knowledge Graph Construction} methodology to build a domain-specific ChipKG. It introduces Circuit Semantic Anchors (CSAs) and the Hierarchical Triple Extraction schema to capture design intent and intricate inter-entity relationships  missed by generic methods.
    
    \item  We introduce the \textbf{ChipKG-Augmented Reasoning} mechanism featuring information-theoretic adaptive retrieval and the CSA-guided semantic filtering layer to enhance the completeness and precision of cross-section dependency tracing.
    
    \item  We build \textbf{SpecEval-QA}, the first benchmark for industrial-scale specification reasoning, incorporating our newly proposed metric, \textbf{Atomic-ROUGE}, robust to paraphrasing and expression variance, thereby aligning evaluation outcomes more closely with human expert judgment.

    % \item We provide a publicly accessible, web-based \textbf{ChipMind} interface(http://116.169.116.27:30011/), accompanied by a tutorial in Appendix A.
    
\end{enumerate}

\section{Related Work}
\subsection{LLMs in Chip Design and Verification}
Large Language Models (LLMs) are increasingly being applied to automate IC development~\cite{chang2024natural,fu2023gpt4aigchip,xu2025revolution}. Seminal works have demonstrated feasibility in generating HDL and HLS code ~\cite{pei2024bettervcontrolledveriloggeneration, wang2025large, wong2024vgv, niu2025rechisel, li2025chathls, ye2025chatmodel, wan2025genben}, creating SystemVerilog Assertions or testbenches ~\cite{ye2025concept, hu2024uvllmautomateduniversalrtl}, and debugging ~\cite{xu2024meic, wang2025veridebug, yao2024location, wan2025fixme}. However, these methods typically operate on isolated, self-contained code snippets and have limited capability to comprehend the cross-document dependencies inherent in industrial-scale circuit specifications. In contrast, our work is specifically designed to enable reasoning across the entire design documentation.

\begin{figure*}[tb]
\centering
\includegraphics[width=\textwidth]{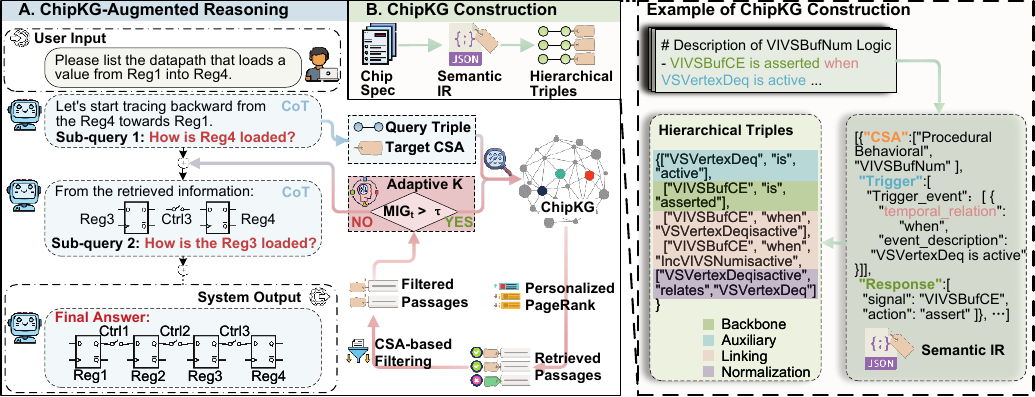}
\caption{Overview of the ChipMind Framework and an Example of ChipKG Construction}
\label{pic:Framework}
\end{figure*}

\subsection{Retrieval-Augmented Generation (RAG)}
RAG is the standard paradigm for grounding LLMs in external documents, but existing methodologies are ill-suited for the highly relational nature of circuit specifications.

\textbf{Standard RAG.} Conventional RAG, whether based on sparse (e.g., BM25~\cite{robertson2009probabilistic,trotman2014improvements}) or dense retrieval~\cite{zhao2024dense,zhan2021optimizing}, relies on semantic similarity. This approach is effective for single-hop, fact-based queries but fundamentally fails to model the multi-hop logical dependencies in hardware design. Consequently, it often retrieves a mixture of relevant but functionally incongruous information, derailing the reasoning process.

\textbf{Knowledge Graph-Augmented RAG (KG-RAG).} To imbue retrieval with structural awareness, recent works augment RAG with KGs. However, existing methods still lack the requisite precision for the IC domain. For instance, GraphRAG~\cite{edge2024local,han2024retrieval} relies on LLM-generated summaries, which risks abstracting away the fine-grained details essential for technical accuracy. To address this loss of granularity, HippoRAG 2~\cite{gutierrez2025rag} innovatively incorporates raw document chunks as nodes directly into the graph. However, its reliance on generic OpenIE techniques yields entity triples that are too coarse to model the complex relationships between entities in circuit specifications. These collective limitations underscore the need for a new KG-RAG paradigm. To this end, we introduce a framework tailored to the unique semantics of chip design, one that constructs a semantically rich and structurally deep knowledge graph.

\subsection{Reasoning-Augmented Frameworks}
Frameworks like Chain-of-Thought (CoT)~\cite{wei2022chain,lyu2023faithful} and Tree-of-Thought (ToT) ~\cite{yao2023tree,long2023large} improve LLM reasoning by decomposing problems into intermediate steps. Integrated approaches such as IRCoT~\cite{trivedi2023interleavingretrievalchainofthoughtreasoning} and ReAct~\cite{yao2023reactsynergizingreasoningacting} incorporate retrieval into this process. Their efficacy, however, is fundamentally bottlenecked by the quality of retrieval at each step. Using generic vector-based RAG or web search APIs is ineffective for the specialized and proprietary nature of design documents, as established above. More critically, advanced ToT frameworks like RATT~\cite{zhang2025ratt} require the reasoning paths to be explicitly predefined. This is incompatible with chip development, where crucial dependencies are often implicit and must be dynamically discovered through iterative exploration. Therefore, our framework departs from these static approaches by introducing a dynamic reasoning process where retrieval and reasoning are synergistically intertwined, allowing the model to actively uncover latent dependencies on-the-fly.

\section{Methodology}
\subsection{ChipMind Framework Overview}

To enable deep, multi-hop reasoning over complex Integrated Circuit (IC) design specifications, we propose \textbf{ChipMind}, a two-stage framework (shown in Figure \ref{pic:Framework}): 

In the first stage, the Circuit Semantic-Aware Knowledge Graph construction methodology is employed. Specifically, semantic content is transformed into a semantic intermediate representation (IR) by functional categorization, and Circuit Semantic Anchors (CSAs) are extracted to identify design intents. Based on semantic IR, structured triples are derived through the Hierarchical Triple Extraction schema, forming a domain-specific Chip Knowledge Graph (ChipKG).

In the second stage, reasoning proceeds through dynamic interactions with ChipKG. When encountering incomplete knowledge during reasoning, the system generates targeted sub-queries, retrieves relevant knowledge graph nodes via vector-similarity matching, and refines their relevance through Personalized PageRank (PPR). Retrieval precision and completeness are further enhanced by employing the adaptive Top-\texttt{K} retrieval strategy based on marginal information gain, along with a CSA-guided semantic filtering layer to eliminate semantic noise from lexically similar yet irrelevant nodes. This structured integration of knowledge and adaptive reasoning systematically decomposes intricate IC reasoning tasks into transparent, traceable steps.

\subsection{Circuit Semantic-Aware Knowledge Graph}
Accurate reasoning over chip specifications requires precise modeling of chip design semantics. However, as discussed in the introduction, generic methods fall short in capturing complex, hierarchical semantic relations within IC documents. Therefore, we propose the \textit{Circuit Semantic-Aware Knowledge Graph Construction} methodology, which explicitly deconstructs and reconnects semantic relationships within circuit documents via a tailored paradigm, enabling comprehensive and robust IC domain-specific knowledge graph (ChipKG) construction. A detailed illustration of this methodology is provided in Figure \ref{pic:Framework}.

\subsubsection{Circuit Semantic Anchoring and Categorization.}
We observe that sentences in IC specifications serve two primary functions:

\begin{itemize}
    \item \textbf{Declarative Functional Description:} Describes static properties and definitions, such as module composition, register fields, and signal functions.
    \item \textbf{Procedural Behavioral Description:} Delineates dynamic logic and behavior, such as state transitions, conditional triggers, and signal assignments.
\end{itemize}

Based on this classification, we utilize specialized parsing templates: declarative descriptions are parsed into central entities and their attributes, while procedural descriptions yield structured “trigger-condition-action” logic. Each sentence is first deconstructed into a concise JSON-based semantic intermediate representation (IR) and distilled into a Circuit Semantic Anchor (CSA). While the IR provides fine-grained details for subsequent triple extraction, the CSA acts as a high-level semantic filter to facilitate efficient downstream reasoning.

\subsubsection{Hierarchical Triple Extraction.}

Based on the extracted IR results, we propose the hierarchical triple extraction schema to comprehensively extract four distinct categories of triples (described below), enabling rich entity relationships within circuit specifications to be restructured into ChipKG.

\begin{itemize}
    \item \textbf{Backbone Triples ($T_B$):} Encode the central action or definition of an entity.
    \item \textbf{Auxiliary Triples ($T_A$):} Encode conditional or temporal semantics that qualify the backbone action.
    \item \textbf{Linking Triples ($T_L$):} Connect $T_B$ with its corresponding $T_A$s, formally representing the dependency.
    \item \textbf{Normalization Triples ($T_N$):} Map compound entities to canonical forms, resolving fragmentation and enhancing structural connectivity.
\end{itemize}

\subsection{ChipKG-Augmented Reasoning}
To trace complex multi-hop dependencies inherent in chip specifications, we propose \textit{ChipKG-Augmented Reasoning}, enabling the LLM to iteratively query ChipKG. This dynamic interaction is driven by two novel components: (1) Adaptive Top-\texttt{K} Retrieval for comprehensive evidence collection, and (2) Semantic Anchor-Guided Filtering for precise noise reduction. This synergy improves the signal-to-noise ratio for more reliable and interpretable reasoning.

\subsubsection{Information-Theoretic Adaptive Top-K Retrieval.} As discussed in Section 1, a fundamental challenge in RAG is the static retrieval number, $K$: a small $K$ misses crucial information, while a large $K$ introduces noise. We thus propose adaptively expanding the retrieval set until marginal utility diminishes.

\textbf{Formalism.} We model the LLM's belief state as a posterior probability distribution $P(A|C_t)$ over the answer space $A$, conditioned on context $C_t=(Q,S_t)$, where $Q$ is the query and $S_t$ is the current document set. The value of new information $\Delta S$, is measured by its impact on this belief state. We quantify this impact using the Kullback-Leibler (KL) divergence and define the Marginal Information Gain $(MIG)$ as:
\[
    MIG(\Delta S \mid \mathcal{C}_t) := D_{\mathrm{KL}}\left( P(A \mid \mathcal{C}_t \cup \Delta S) \, \big\| \, P(A \mid \mathcal{C}_t) \right)
\]

A significant $MIG$ indicates that $\Delta S$ provides novel, decision-relevant evidence, whereas a near-zero value implies redundancy and signals retrieval termination.

\begin{algorithm}[tb] 
\small
\caption{Iterative Context Expansion Algorithm}
\label{alg:context_expansion}
\begin{algorithmic}[1] % [1] 表示显示行号
    \STATE \textbf{Initialize:} Form initial set $S_0$ by retrieving top-$k_0$ documents.
    \FOR{$t = 0, 1, 2, \dots$}
        \STATE \textbf{Expand:} Retrieve next $\Delta k$ documents to form incremental set $\Delta S_t$.
        \STATE \textbf{Estimate Belief State (Proxy for intractable $P(A|C)$):}
        \STATE \quad Generate summary $A'_{\text{base}} \leftarrow \text{LLM}(S_t)$.
        \STATE \quad Generate summary $A'_{\text{new}} \leftarrow \text{LLM}(S_t \cup \Delta S_t)$.
        \STATE \quad Calculate gain $MIG_t$ based on the change between embeddings of $A'_{\text{base}}$ and $A'_{\text{new}}$.
        
        \STATE \textbf{Check \& Terminate:}
        \IF{$MIG_t > \tau$}
            \STATE $S_{t+1} \leftarrow S_t \cup \Delta S_t$. \COMMENT{Information is useful, continue.}
        \ELSE
            \STATE \textbf{return} $S_t$. \COMMENT{Diminishing returns, terminate.}
        \ENDIF
    \ENDFOR
\end{algorithmic}
\end{algorithm}

\textbf{Iterative Retrieval Algorithm.} As detailed in Algorithm~\ref{alg:context_expansion}, the context set $S$ is iteratively expanded with evidence nodes retrieved from the ChipKG. Since directly computing $P(A|C)$ is infeasible, we approximate Marginal Information Gain (MIG) by prompting the LLM to summarize contexts with ($A'_{\text{new}}$) and without ($A'_{\text{base}}$) new nodes. The MIG at each iteration is then estimated as the cosine distance between their embeddings:
\[
    MIG_t \approx 1- \cos{(emb(A'_{base}), emb(A'_{new}))}
\]
We terminate the retrieval once $MIG_t$ drops below a threshold $\tau$, indicating diminishing returns.

\subsubsection{CSA-based Semantic Filtering.} 
While adaptive retrieval ensures evidence completeness, it can also introduce functionally irrelevant information. To maintain retrieval precision, we introduce a semantic filtering layer guided by Circuit Semantic Anchors (CSAs). Specifically, for each query or sub-task $Q$, we prompt the LLM to infer the query's intent by generating a target anchor $CSA_{\text{target}} = (\text{type}_{\text{target}}, \text{entity}_{\text{target}})$.

The candidate set $S_{cand}$ is pruned to form the final context $S_{final}$ by retaining only nodes whose anchors exactly match the target anchor $CSA_{\text{target}}$:
\[
    S_{final} = \{s_i \in S_{cand} \mid \text{Compat}(CSA_i, CSA_{\text{target}})\}
\]

This precise filtering removes irrelevant information, enhancing signal-to-noise ratio for downstream reasoning.

\subsubsection{ChipKG-Augmented Reasoning Workflow.}
We integrate the preceding components into our ChipKG-Augmented Reasoning framework (Algorithm \ref{alg:CAR}).

An iterative loop begins by reasoning on the current context and introspectively detecting knowledge gaps. Once a gap is identified, ChipMind formulates a targeted sub-query, retrieves relevant evidence from ChipKG, and integrates refined evidence back into reasoning. This loop repeats until sufficient context is collected to synthesize a final, grounded answer.

\begin{algorithm}[tb]
\small
\caption{ChipKG-Augmented Reasoning Workflow}
\label{alg:CAR}
\textbf{Input}: An initial query $Q_{in}$ \\
\textbf{Output}: A final, grounded answer $A_{final}$ \\
\begin{algorithmic}[1] %[1] enables line numbers
\STATE Initialize: $C \leftarrow Q_{in}$; $A_{final} \leftarrow \text{null}$;
\WHILE{termination condition not met}
\STATE // \textbf{Step 1: Reason and Detect Knowledge Gaps}
\STATE $r_t \leftarrow \text{Reason}(C)$;
\IF{\text{DetectUncertainty}$(r_t, C)$}
\STATE \hspace{-1.5em} // \textbf{Step 2: Formulate Information Need}
\STATE $(q_t, CSA_{target}) \leftarrow \text{Formulate-Sub-Query}(r_t)$;
\STATE \hspace{-1.5em} // \textbf{Step 3: Acquire \& Refine Knowledge}
\STATE $S_{cand} \leftarrow \text{AdaptiveKRetrieve}(q_t, \text{ChipKG})$;
\STATE $S_{filtered} \leftarrow \text{CSAGuidedFilter}(S_{cand}, CSA_{target})$;
\STATE \hspace{-1.5em} // \textbf{Step 4: Integrate and Continue Reasoning}
\STATE $C \leftarrow C \cup S_{filtered}$;
\ELSE
\STATE \hspace{-1.5em} // \textbf{Step 5: Synthesize and Terminate}
\STATE $A_{final} \leftarrow \text{Synthesize-Answer}(C)$;
\STATE \textbf{break};
\ENDIF
\ENDWHILE
\STATE \textbf{return} $A_{final}$
\end{algorithmic}
\end{algorithm}

Our framework advances existing multi-step retrieval methods in two key aspects: we replace static, fixed-\texttt{K} retrieval with a dynamic, information-theoretic strategy to optimize context size, and elevate filtering from shallow semantic matching to intent-level precision using CSA-guided functional alignment. This synergy improves the signal-to-noise ratio for more reliable and interpretable reasoning.

\subsection{Atomic-ROUGE}
\label{sec:atomic_rouge}

Traditional n-gram overlap metrics (e.g., ROUGE) struggle with semantic variations, while modern metrics like BERTScore improve semantic matching but still fail to reliably evaluate factual correctness---often misled by plausible hallucinations or partially correct statements. To overcome this fundamental limitation, we propose Atomic-ROUGE, a novel metric explicitly designed to evaluate factual fidelity in complex reasoning tasks.

The core of our approach is a two-part decomposition: first, human experts decompose the reference answer $y$ into a set of minimal, self-contained semantic ``atomic facts'' $A_{ref} = \{a_1, a_2, ..., a_n\}$. In parallel, a powerful LLM similarly decomposes the generated answer $\hat{y}$ to produce the set $A_{gen}$. An LLM semantic judge then identifies correct matches $A_{matched}$ via semantic equivalence:
    \[
    \begin{aligned}
    A_{matched} = \{ & \hat{a_j} \in A_{gen} \mid \exists a_i \in A_{ref} \\
    & s.t.\ \text{IsSemanticallyEquivalent}(\hat{a_j}, a_i) \}
    \end{aligned}
    \]

Finally, drawing from classic information retrieval, precision, recall, and F1-score quantify factual correctness, completeness, and balanced semantic fidelity, respectively:
    \[
P=\frac{\mid A_{matched}\mid}{\mid A_{gen}\mid}, R=\frac{\mid A_{matched} \mid}{\mid A_{ref}\mid}, F1 = 2 \cdot \frac{P \cdot R}{P+R}
    \]

\section{Evaluation}

To rigorously evaluate ChipMind, we introduce a challenging benchmark specifically designed for industrial-scale Integrated Circuit (IC) specifications. We systematically compare our framework against strong baselines, evaluating performance using our proposed Atomic-ROUGE metric. Furthermore, we also validate the effectiveness and reliability of Atomic-ROUGE itself.

\subsection{Experimental Setup}

\begin{table}[t]
\centering
\small
\renewcommand{\arraystretch}{1.4}
\newcolumntype{M}[1]{>{\raggedright\arraybackslash}m{#1}}
\begin{tabular}{@{}M{0.3\columnwidth}M{0.17\columnwidth}M{0.43\columnwidth}@{}}
\toprule
\textbf{Question Type} & \textbf{\# of Hops} & \textbf{Description} \\ 
\midrule
\makecell[l]{Single-Module\\Config Loc} & 1 & Identify signals/parameters from single-paragraph text. \\
% \midrule
\makecell[l]{Cross-Module\\Config Loc} & 5 $\sim$ 12 & Locate signals/parameters across multiple paragraphs. \\
% \midrule
\makecell[l]{Behavioral Process\\Analysis} & 5 $\sim$ 8 & Reason about internal module procedures. \\
% \midrule
\makecell[l]{Signal Dependency} & 2 $\sim$ 5 & Trace signal flow across modules. \\
% \midrule
Control Path Tracing & 2 & Evaluate FSM transitions and logic. \\
\bottomrule
\end{tabular}
\caption{Question Types and Descriptions for SpecEval-QA}
\label{tab:SpecEval}
\end{table}

\begin{table*}[htbp]
  \centering
  \small 
  \setlength{\tabcolsep}{1.8pt} 
  \renewcommand{\arraystretch}{1.1}
  \newcolumntype{M}[1]{>{\centering\arraybackslash}m{#1}}
  \begin{tabular}{
        ll|
        M{0.9cm}M{0.9cm}|
        M{0.9cm}M{0.9cm}|
        M{0.9cm}M{0.9cm}|
        M{0.9cm}M{0.9cm}|
        M{0.9cm}M{0.9cm}|
        M{0.9cm}
    }
    \hline
    \multicolumn{2}{c|}{\textbf{Question Type}} & 
    \multicolumn{2}{M{1.9cm} |}{\textbf{Signal-Module Config Loc}} &
    \multicolumn{2}{M{1.9cm} |}{\textbf{Cross-Module Config Loc}} &
    \multicolumn{2}{M{1.9cm} |}{\textbf{Behavioral Process Analysis}} & 
    \multicolumn{2}{M{1.9cm} |}{\textbf{Signal Dependency}} & 
    \multicolumn{2}{M{1.9cm} |}{\textbf{Control Path Tracing}} & 
    \multicolumn{1}{M{0.9cm}}{\multirow{2}{*}{\textbf{AVG}}} \\
    \cline{1-12}
    \textbf{Model Type} & \textbf{Model Name} & AVG & STD & AVG & STD & AVG & STD & AVG & STD & AVG & STD \\
    \hline
    % \noalign{\vskip 2pt}
    \multirow{4}[2]{*}{Vector RAG} 
        & GPT-4.1 & 0.82 & 0.15 & 0.63 & 0.26 & 0.87 & 0.04 & 0.85 & 0.21 & 0.78 & 0.17 & \underline{0.79} \\
        & Claude-4 & 0.76 & 0.21 & 0.56 & 0.38 & 0.82 & 0.09 & 0.82 & 0.19 & 0.71 & 0.31 & 0.73 \\
        & Llama-4-scout & 0.93 & 0.15 & 0.49 & 0.25 & 0.60 & 0.22 & 0.60 & 0.19 & 0.60 & 0.32 & 0.64 \\
        & DeepSeek-R1 & 0.86 & 0.18 & 0.60 & 0.26 & 0.72 & 0.13 & 0.89 & 0.18 & 0.64 & 0.30 & 0.78 \\
    \hline
    \multirow{3}[2]{*}{KG-based RAG} 
        & GraphRAG & 0.56 & 0.52 & 0.45 & 0.48 & 0.76 & 0.11 & 0.40 & 0.42 & 0.77 & 0.17 & 0.55 \\
        & HippoRAG 2 & 0.94 & 0.13 & 0.73 & 0.25 & 0.78 & 0.19 & 0.68 & 0.35 & 0.72 & 0.18 & \underline{0.76} \\
        & LightRAG & 0.68 & 0.42 & 0.38 & 0.15 & 0.69 & 0.20 & 0.63 & 0.21 & 0.55 & 0.18 & 0.61 \\
    \hline
    \multirow{3}[2]{*}{\makecell[l]{Reasoning-augmented\\Framework}} 
        & ReAct & 0.94 & 0.13 & 0.67 & 0.16 & 0.73 & 0.12 & 0.72 & 0.30 & 0.90 & 0.11 & 0.79 \\
        & IRCoT & 0.93 & 0.15 & 0.59 & 0.21 & 0.88 & 0.12 & 0.83 & 0.18 & 0.79 & 0.16 & \underline{0.81} \\
        & Search-o1 & 0.76 & 0.43 & 0.59 & 0.41 & 0.68 & 0.24 & 0.71 & 0.29 & 0.74 & 0.17 & 0.70 \\
    \hline
    \makecell[l]{\textbf{KG-augmented}\\\textbf{Reasoning}} & \textbf{ChipMind} & \textbf{1.00} & \textbf{0.00} & \textbf{0.84} & \textbf{0.15} & \textbf{0.93} & \textbf{0.02} & \textbf{0.97} & \textbf{0.06} & \textbf{1.00} & \textbf{0.00} & \textbf{0.95} \\
    \hline
  \end{tabular}
  \caption{Performance Comparison (Atomic-ROUGE F1 Scores) of ChipMind and Baseline Methods}
  \label{tab:overall performance}
\end{table*}

\subsubsection{Benchmark.} Current benchmarks for LLM-aided hardware design (LAD) fall short of reflecting the scale and complexity of real-world chip specifications. To address this limitation, we propose \textbf{SpecEval-QA}, a novel question-answering benchmark derived from a comprehensive, industrial-grade specification of a High-Performance Compute and Interconnect Macro-block (51k tokens).

SpecEval-QA features 25 questions across representative, realistic chip-level reasoning scenarios (Table \ref{tab:SpecEval}) designed to test long circuit specification comprehension and reasoning over 1$\sim$12 reasoning hops. Critically, each question is rigorously annotated with a gold answer, associated atomic facts, and corresponding ground-truth supporting passages. Such detailed annotations enable a fine-grained verification of factual correctness and reasoning pathways, forming the basis for our new metrics.

\subsubsection{Baselines.} Since no existing end-to-end frameworks directly address this task, we comprehensively evaluate ChipMind against three categories of strong baselines:
\begin{itemize}
    \item \textbf{Vector RAG:} The standard vector-based retrieval methods (BGE-M3~\cite{chen2024bge}) with state-of-the-art (SOTA) LLMs (see Model Selection and Fairness for details), representing the conventional RAG approach.
    \item \textbf{KG-RAG:} Advanced KG-enhanced methods, including GraphRAG~\cite{edge2024local}, HippoRAG 2~\cite{gutierrez2025rag}, and LightRAG~\cite{guo2024lightrag}, to assess our domain-specific ChipKG against generic graph-based solutions.
    \item \textbf{Reasoning-Augmented Framework:} Advanced reasoning-driven approaches like ReAct~\cite{yao2023reactsynergizingreasoningacting}, IRCoT~\cite{trivedi2023interleavingretrievalchainofthoughtreasoning}, and Search-o1~\cite{li2025search}, specifically benchmarking the effectiveness of ChipMind's KG-augmented reasoning against contemporary reasoning strategies.
\end{itemize}

\subsubsection{Model Selection and Fairness.} We select several top-tier LLMs, including: SOTA closed-source models, GPT-4.1~\cite{openai_gpt4_1_2025} and Claude-4~\cite{anthropic_claude4_2025}; a strong open-source baseline Llama-4-scout~\cite{meta_llama4_2025}; and a reasoning-optimized open-source model DeepSeek-R1~\cite{guo2025deepseek}. Because experimental results (Table \ref{tab:overall performance}) show that GPT-4.1 achieves the highest performance, we adopt it as the backbone across all KG-RAG and reasoning-augmented frameworks. This standardization ensures observed performance differences reflect framework design rather than LLM capability. The sole exception is Search-o1, whose reasoning process is tightly coupled with its native QwQ-32B-Preview~\cite{qwen_team_qwq_2024} model. Additionally, to guarantee fairness in retrieval, all retrieval-dependent frameworks utilize BGE-M3 retriever, accessing an identical data corpus.

\subsubsection{Prototyping and Implementation.} ChipMind is designed as a model-agnostic framework, compatible with various capable LLMs. In our implementation, we selected DeepSeek-R1 as the core reasoning engine, given its proven effectiveness in logic-intensive tasks and open accessibility that ensures reproducibility. Tasks demanding strong instruction-following capabilities, such as ChipKG construction and Atomic-ROUGE evaluation, leveraged GPT-4.1. We set the sampling temperature to 0.7 for generative tasks and 0.2 for evaluation to improve stability and reproducibility. All experiments were performed on a server with dual Intel Xeon Platinum 8480+ CPUs and eight NVIDIA H20 GPUs (96 GB each).

\begin{figure}[t]
\centering
\includegraphics[width=\columnwidth]{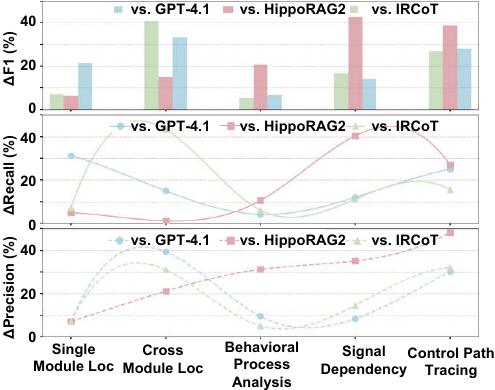} 
\caption{Performance Gains by Task Type}
\label{fig:performance_improvement}
\end{figure}

\subsection{End-to-End Performance}

\subsubsection{Evaluation Metric.} As detailed in Section \ref{sec:atomic_rouge}, we adopt our proposed Atomic-ROUGE metric for evaluation. By measuring semantic overlap between atomic facts extracted from generated and reference answers, Atomic-ROUGE robustly quantifies precision, recall, and F1-score, effectively handling semantic variations and paraphrasing. We further validate Atomic-ROUGE itself in Section \ref{sec:atomic-rouge-validation}.

\subsubsection{Overall Performance.} To ensure robust evaluation, each method was executed 5 times, with each run assessed 20 times using Atomic-ROUGE. Outliers were removed via the $2\sigma$ rule before averaging. As summarized in Table~\ref{tab:overall performance}, ChipMind achieves a SOTA mean F1-score of 0.95, outperforming all baselines by an average of 34.59\% and a maximum gain of 72.73\% compared to GraphRAG. ChipMind consistently surpasses the strongest baselines within each category, with improvements of 20.25\% over GPT-4.1 (F1=0.79), 25\% over HippoRAG 2 (F1=0.76), and 17.28\% over IRCoT (F1=0.81).
% Detailed per-question scores are provided in Appendix C.

\subsubsection{Performance by Task Type.}

\begin{figure}[t]
\centering
\includegraphics[width=\columnwidth]{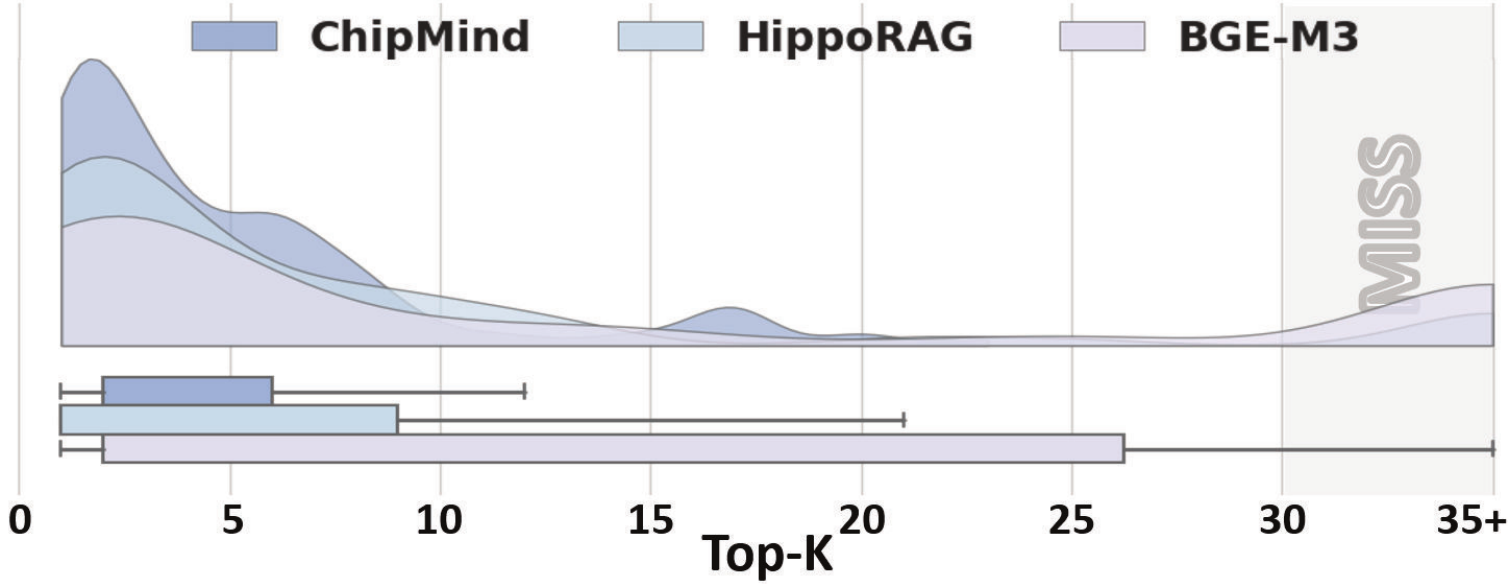} 
% \caption{Retrieval Distribution at Top-\texttt{K}: Half-violin plots (top) and box-and-whisker plots (bottom).}
\caption{Retrieval Distribution at Top-\texttt{K}. 
Top: Half-violin plots illustrating probability density. 
Bottom: Box-and-whisker plots summarizing distribution characteristics.}
\label{pic:CloudRain}
\end{figure}

Figure \ref{fig:performance_improvement} compares ChipMind’s performance improvements across task types against top baselines (GPT-4.1, HippoRAG 2, and IRCoT), highlighting component-level advantages:

For single and cross-module configuration localization tasks, ChipMind substantially outperforms GPT-4.1 and IRCoT, which rely on unstructured RAG. This advantage stems from ChipMind’s KG-enhanced retrieval for greater precision, coupled with adaptive iterative retrieval that ensures comprehensive context coverage and improved recall.

For complex reasoning tasks (behavioral processes, signal dependencies, control paths), ChipMind markedly surpasses HippoRAG 2's single-round static retrieval by decomposing tasks into iterative, verifiable reasoning steps, capturing deep dependencies overlooked by single-step retrieval.

\begin{figure}[t]
\centering
\includegraphics[width=\columnwidth]{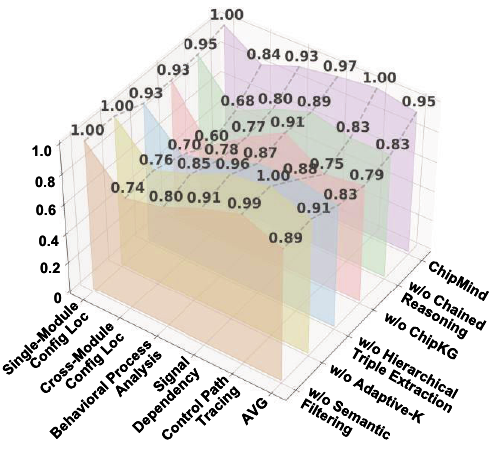} 
\caption{Results of Ablation Experiments}
\label{pic:ablation}
\end{figure}

\subsection{Retrieval Performance Analysis}
\label{sec:retrieval_performance}
To isolate the source of ChipMind’s performance gains, we compare the retrieval completeness of vector-based RAG (BGE-M3), KG-RAG (HippoRAG 2), and ChipMind using \textbf{System Recall@$K$}, defined as the fraction of ground-truth passages retrieved during the entire reasoning process.

As shown in Fig.~\ref{pic:CloudRain}, ChipMind achieves near-perfect System Recall@20 of 99.2\%, significantly outperforming single-pass methods HippoRAG 2 (86.8\%) and BGE-M3 (70.5\%). This confirms that the baselines' lower performance primarily stems from incomplete evidence retrieval.

The performance gap arises from semantic mismatches in multi-hop queries, as initial queries are often semantically distant from intermediate evidence, causing single-pass retrieval to rank critical passages poorly. ChipMind bridges this gap via iterative sub-query generation, progressively aligning with intermediate evidence and elevating essential passages to top-ranked positions for complete retrieval.

\subsection{Atomic-ROUGE Validation}
\label{sec:atomic-rouge-validation}

To validate that Atomic-ROUGE aligns with human judgments, we compared its scores against expert ratings. Three senior chip engineers independently rated ChipMind outputs on 10-point scales for Semantic Fidelity and Answer Completeness; ratings were averaged to form the final human score. Atomic-ROUGE achieved a Pearson correlation of 0.83 with human ratings, surpassing the BERTScore ($r=0.71$). This confirms Atomic-ROUGE as a reliable automated metric for expert evaluation in fact-intensive tasks.

\subsection{Ablation Study}

To quantify the contribution of each key component, we conducted four ablation studies, systematically disabling modules from the full ChipMind framework (Figure ~\ref{pic:ablation}).

\subsubsection{Contribution of ChipKG-Augmented Reasoning.}

Replacing the multi-turn loop with single-pass reasoning sharply decreases performance on multi-hop tasks, confirming the necessity of dynamic sub-query generation for complete evidence retrieval (Section \ref{sec:retrieval_performance}). Notably, even the degraded single-pass variant (F1=0.83) surpasses KG-RAG baselines such as HippoRAG 2 (0.76), underscoring the inherent schema advantage of our domain-specific ChipKG.

\subsubsection{Contribution of ChipKG.}
Replacing ChipKG with standard Vector RAG severely reduces precision on localization tasks. Further substituting our triple-extraction schema with OpenIE, although outperforming Vector RAG, remains inferior to the full system. This highlights our ChipKG schema’s crucial role in modeling the logic of chip specifications.

\subsubsection{Contribution of Adaptive-K Retrieval.} The adaptive-$K$ mechanism is the most critical for cross-module configuration localization tasks, as dynamically expanding the search scope ensures comprehensive information coverage.

\begin{figure}[t]
\centering
\includegraphics[width=\columnwidth]{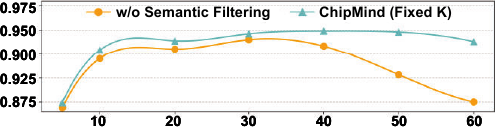} 
\caption{Effect of Semantic Filtering vs. Top-\texttt{K}}
\label{pic:K-effect}
\end{figure}

\subsubsection{Contribution of Semantic Filtering.} 
The CSA-guided filter mitigates context pollution, with performance sharply degrading as $K$ increases without it, due to noise from irrelevant documents (Figure \ref{pic:K-effect}). In contrast, ChipMind maintains robust performance even at $K=50$, highlighting the filter’s role in preserving signal-to-noise ratios and supporting larger retrieval budgets.

% The CSA-guided filter is essential to mitigate context pollution. We use the fixed $K$ value and find that ablating this filter sharply degrades performance as $K$ increases, due to noise from irrelevant documents, as shown in Figure \ref{pic:K-effect}. In contrast, the full ChipMind maintains robust results even at $K=50$, validating the filter’s role in preserving high signal-to-noise ratios and enabling larger retrieval budgets.

\section{Conclusion}

This paper introduced ChipMind, a graph-enhanced reasoning framework designed for the unique complexities of hardware specifications. ChipMind deconstructs implicit logic from specification text into an explicit Chip Knowledge Graph (ChipKG), and its graph-augmented engine enables an LLM to perform verifiable, multi-hop reasoning with high contextual sufficiency and precision. This work sheds light on how structured reasoning can enhance the viability of LLMs for industrial LLM-aided hardware design challenges.

\section*{Acknowledgments}

This work is supported by the National Key Research and Development Program of China (Grant No. 2024YFB4405600) and the National Natural Science Foundation of China (Grant No. 92464301). We gratefully acknowledge the support from these programs.

\bibliography{aaai2026}

\end{document}